\documentclass[runningheads]{llncs}
\usepackage[T1]{fontenc}
\usepackage{graphicx}
\usepackage{cite}
\usepackage{amssymb,amsfonts}
\usepackage{algorithmic}
\usepackage{graphicx}
\usepackage{textcomp}
\usepackage{xcolor}
\usepackage{multirow}
\usepackage{array}
\usepackage{comment}
\usepackage{booktabs}
\usepackage{bibentry}
\usepackage{subcaption}

\usepackage{color}

\title{Can GPT-4 Help Detect Quit Vaping Intentions? An Exploration of Automatic Data Annotation Approach}
\titlerunning{Can GPT-4 Help Detect Quit Vaping Intentions?}
\author{
Sai Krishna Revanth Vuruma\inst{1}\orcidID{0009-0009-3741-9343} \and
Dezhi Wu\inst{1}\orcidID{0000-0002-3554-1136} \and
Saborny Sen Gupta\inst{1}\orcidID{0009-0008-0937-5366} \and
Lucas Aust\inst{1} \and
Valerie Lookingbill\inst{1}\orcidID{0000-0003-1453-2633} \and
Wyatt Bellamy\inst{1}\orcidID{0009-0000-4616-9715} \and
Yang Ren\inst{1}\orcidID{0000-0002-6128-5826} \and
Erin Kasson\inst{2} \and
Li-Shiun Chen\inst{2}\orcidID{0000-0001-6762-5054} \and
Patricia Cavazos-Rehg\inst{2}\orcidID{0000-0003-3352-1198} \and
Dian Hu\inst{3}\orcidID{0000-0003-1277-142X} \and
Ming Huang\inst{3}\orcidID{0000-0001-7367-3626}
}
\authorrunning{S. Vuruma et al.}
\institute{
HI3 Tech Lab, University of South Carolina, Columbia SC, USA
\email{dezhiwu@cec.sc.edu}\\
\and
School of Medicine, Washington University, St. Louis MO, USA
\and
Center of Translational AI Excellence and Applications in Medicine, University of Texas Health Science Center, Houston TX, USA\\
\email{ming.huang@uth.tmc.edu}
}

\begin{document}
\maketitle              % typeset the header of the contribution
\begin{abstract}
In recent years, the United States has witnessed a significant surge in the popularity of vaping or e-cigarette use, leading to a notable rise in cases of e-cigarette and vaping use-associated lung injury (EVALI) that caused hospitalizations and fatalities during the EVALI outbreak in 2019, highlighting the urgency to comprehend vaping behaviors and develop effective strategies for cessation. Due to the ubiquity of social media platforms, over 4.7 billion users worldwide use them for connectivity, communications, news, and entertainment with a significant portion of the discourse related to health, thereby establishing social media data as an invaluable organic data resource for public health research. In this study, we extracted a sample dataset from one vaping sub-community on Reddit to analyze users' quit-vaping intentions. Leveraging OpenAI's latest large language model GPT-4 for sentence-level quit vaping intention detection, this study compares the outcomes of this model against layman and clinical expert annotations. Using different prompting strategies such as zero-shot, one-shot, few-shot and chain-of-thought prompting, we developed 8 prompts with varying levels of detail to explain the task to GPT-4 and also evaluated the performance of the strategies against each other. These preliminary findings
emphasize the potential of GPT-4 in social media data analysis, especially in identifying users’ subtle intentions that may elude human detection.

\end{abstract}

\section{Introduction}
Studies indicate that epidemic levels of consumption was observed among adolescents and young adults during the last decade, with a massive increase in sale and usage of e-cigarettes and other disposable vaping products \cite{dai_prevalence, stalgaitis2020vaping} leading to the EVALI outbreak in 2019. With the nation's youth emerging as the high risk population, research suggests that many e-cigarette users are unaware of the potential dangers of vaping such as Vape Dependence\cite{soule2020cannot} and Stealth-vaping\cite{yingst2019cigarette}, with vape frequency directly associated with perceived satisfaction while being indirectly associated with perceived danger\cite{kozlowski2017daily}. Vaping products contain cancer-causing agents, toxins, heavy metals, and other harmful particles that are substantially linked to lung, heart, and brain damage \cite{mckay2024assessing}. Recent efforts towards educating the young populace about the negative impacts of vaping have resulted in a large number of e-cigarette users intending to quit vaping \cite{struik2021cigarette}, with about 45\% of young vapers reporting interest in quitting, while 25\% attempted to quit in 2020-2021 \cite{smith2021intention}. The goal is to now identify these users and help them quit vaping by proving the necessary resources to support them along the way.

Contemporary research studies have leveraged popular social media platforms such as Twitter and Reddit for public surveillance of health topics. Approximately, more than 70\% of people use at least one social media platform and the number of new users in any of these popular platforms is increasing everyday especially among users aged 18-29\cite{kwon2020perceptions}. The utilization of social media data emerges as a nascent source of public health information, offering novel insights into public health trends and enhancing the capabilities of public health surveillance.

Previous vaping studies such as \cite{wu2022vapetwitter, KASSON2021104574} used topic modeling and sentiment analysis along with clinical insights on social media data to show users on these platforms might benefit from digital intervention programs for vaping cessation. For clinicians to potentially employ proactive outreach strategies to engage vaping patients for education and treatment on social media platforms, it is imperative to conduct further research into the analysis of vaping discourse on these platforms \cite{ketonen2020characterizing}, aiming to develop Artificial Intelligence (AI)-based approaches to more efficiently and accurately identify these users' vaping behaviors and develop targeted vaping prevention and intervention programs for the youth population.

In this preliminary study,  we aim to employ and evaluate OpenAI's GPT-4 model against layman and clinical expert annotators on a sentence-level annotation task to identify vaping cessation interests among Reddit users. Our preliminary findings indicate that the GPT-4 model performs impressively, but it still has a ways to go before replacing human annotators.
\section{Literature Review}
Interpretation of natural language data extracted from social media platforms requires deep contextual knowledge and understanding, lack of which can lead to incorrect labeling and annotations \cite{liyanage2023gpt}. Manual annotations of these texts can be challenging for humans as they are often short, informal and contain different socio-cultural opinions and perceptions \cite{maceda2023classifying}. Care must be taken while using state-of-the-art Machine Learning (ML) algorithms and Natural Language Processing techniques in tasks requiring complex inferences as shown in \cite{tornberg2023chatgpt, cheng2023gpt}. Traditional ML and Deep Learning models like CNNs, RNNs and pretrained Language Models like BERT require a high-quality annotated corpus to develop an effective model for sentiment analysis. On the other hand, advanced and intuitive Large Language Models (LLMs) such as OpenAI’s Generative Pre-Trained Transformer models GPT-3 \cite{brown2020language} and GPT-4 \cite{achiam2023gpt} among others, allow zero-shot learning, one-shot learning, and few-shot learning which could be used for detecting quit vaping intention without intensive training. These LLMs have shown proficiency in in-context learning where they outperformed traditional methods \cite{alhamed2024using, deng2023llms} and can generate quick results while not being susceptible to some of the limitations observed in human annotation \cite{pangakis2023automated}.

On Data Annotation tasks, studies have shown that ChatGPT's performance is promising in classifying and generating explanations for implicit sentiment analysis such as hate speech detection \cite{huang2023chatgpt}, zero-shot sentence-level annotation of legal documents \cite{savelka2023unlocking}, political tweet labeling \cite{tornberg2023chatgpt} and identifying adverse events about a cannabis-derived product \cite{leas2024using}. Although works such as \cite{ding2022gpt} reiterates ChatGPT is emerging as a potential alternative to human annotation as it is faster and cheaper, some researchers advise caution and argue that human-in-the-loop validation must be maintained to guarantee the reliability of its results \cite{thapa2023humans}. In contrast, ChatGPT-generated Natural Language Explanations (NLEs) can influence human perceptions and can result in a risk of misleading common people in case of incorrect predictions \cite{huang2023chatgpt}.

OpenAI's GPT-4 model has shown remarkable capabilities in a multitude of domains, even clearing the bar exam according to a recent study\cite{katz2024gpt}. Research indicates that GPT-4 can act as an alternative to layman annotation in many diverse areas\cite{tornberg2023chatgpt, cheng2023gpt}.
\section{Methodology}
The workflow adopted for this study is illustrated in Figure \ref{fig_workflow}. Each stage of the pipeline will be discussed in detail in the respective subsections. First the data is extracted from Reddit and cleaned, then it is sent to the annotators: layman, expert and the GPT-4 model for annotation. The performance of all three annotators is compared at the end to draw conclusions. With the expert annotated dataset as the ground truth, we will use two types of metrics: qualitative and quantitative for formulating the results.

\subsection{Data Collection \& Preparation}
In the popular social media platform Reddit, r/QuitVaping is the largest subreddit dedicated to help users quit vaping and other tobacco products with around 40,000 subscribers. Using Reddit's Async PRAW API, we extracted a total of 1000 posts from the aforementioned r/QuitVaping subreddit. These posts ranged from users talking about their progress towards quitting vaping to users looking for help or motivation to quit or reduce vape use. Out of these 1000 posts, approximately 120 were randomly selected to form a sample dataset. From each post in the sample dataset, two columns, namely title and body were extracted and broken down into sentences using the Sentence Tokenizer from the NLTK library \cite{nltk_sent_tokenize}. Any sentence that had less than 3 tokens was dropped and so were duplicates. A total of 1059 sentences were available for annotation.

\subsection{Human Annotation}
\subsubsection{Layman Annotation}
Two layman annotators were tasked with labeling the cleaned sentences as 'YES' if the speaker explicitly mentions their idea, desire, decision, plan, or action to quit vaping. And to label them as 'NO' otherwise. For a sentence to be labeled as 'YES', there must be a clear indication that the speaker intends to quit vaping. Discrepancies (n=28) were resolved internally with an Inter-coder Reliability score (ICR) of 0.78.

\subsubsection{Expert Annotation}
Two clinical experts from the School of Medicine, Washington University were asked to perform the expert annotation by following the same guidelines mentioned above. The coders independently reviewed the dataset and coded all the sentences. The second coder resolved discrepancies (n=22).

\subsection{GPT-4 Annotation}
Interaction with the GPT-4 model can be done via prompts that must be carefully constructed to get the best performance out of the model. Each prompt let's you assign a 'role' which indicates who the sender of that message (prompt) is. Taking inspiration from the prompt templates used in \cite{han2023well, zhang2024good} we devised the prompts for our study using approaches like zero-shot, one-shot, few-shot and chain-of-thought prompting.

\begin{figure}[tb]
\begin{subfigure}{0.5\textwidth}
    \includegraphics[width=\textwidth, height=4cm]{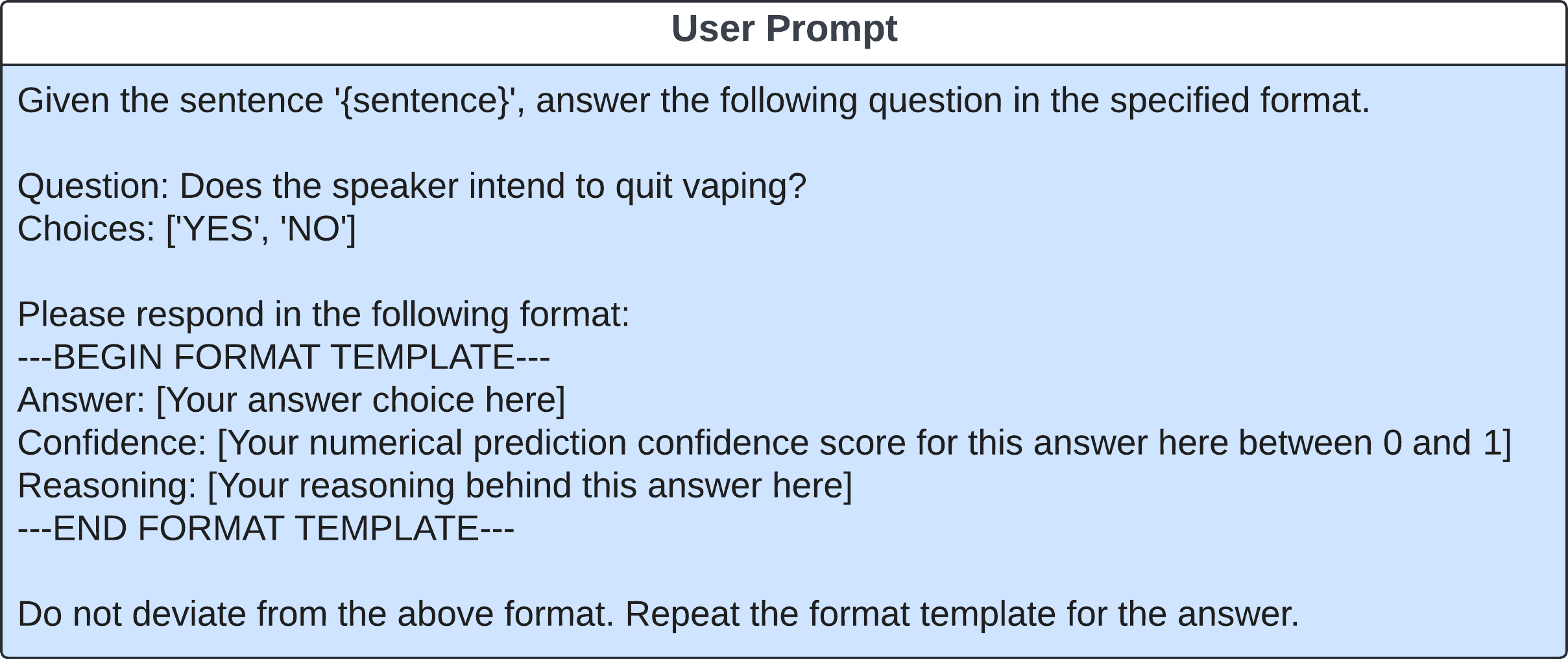}
    \caption{Low Detail}
    \label{fig_user_prompt_low}
\end{subfigure}
\begin{subfigure}{0.5\textwidth}
    \includegraphics[width=\textwidth, height=4cm]{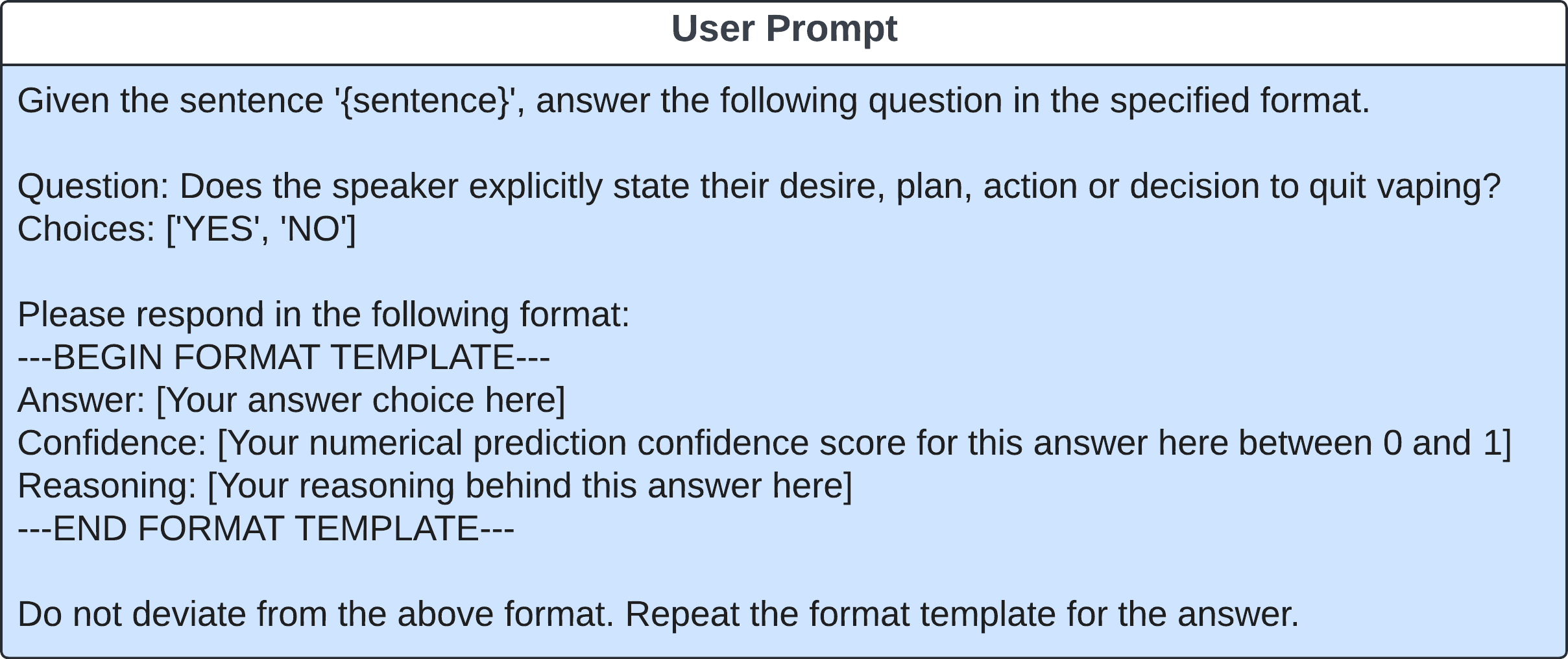}
    \caption{High Detail}
    \label{fig_user_prompt_high}
\end{subfigure}
\caption{Sample User Prompts}
\label{fig_user_prompt}
\end{figure}

Given a sentence, the model was tasked to return a label, a numerical confidence score and its reasoning for choosing that label for that sentence. Figure \ref{fig_sys_prompt} shows the system prompt that we used to introduce the context of the task to the GPT-4 model, while Figure \ref{fig_user_prompt} contains a sample user prompt that passes the input data along with instructions on how the model should respond.

As shown in Table \ref{tab_prompt_list}, we developed 8 prompts using different prompting strategies plus another variable called 'detail'. The low detail prompts (P1-P4) have the structure shown in Figure \ref{fig_user_prompt_low} with the question phrased using simpler language, i.e., "Does the speaker intend to quit vaping?", while the high detail prompts (P5-P8) use a more directed question as shown in Figure \ref{fig_user_prompt_high}. The one-shot and few-shot variants include examples in the user prompt, while the chain-of-thought prompts include the phrase "think step-by-step" in the question.

\begin{table}[tb]
\centering
\caption{Prompts Used. Here, detail column determines how vague (Figure \ref{fig_user_prompt_low}) or specific (Figure \ref{fig_user_prompt_high}) the question is phrased in the user prompt.}
\label{tab_prompt_list}
\begin{tabular}{|c|c|c|}
\hline
\textbf{Prompt ID} &  \textbf{Strategy} & \textbf{Detail}\\
\hline
P1  &   zero-shot   &   low   \\
P2  &   zero-shot, chain-of-thought   &   low   \\
P3  &   one-shot   &   low   \\
P4  &   few-shot   &   low   \\
P5  &   zero-shot   &   high   \\
P6  &   zero-shot, chain-of-thought   &   high   \\
P7  &   one-shot   &   high   \\
P8  &   few-shot   &   high   \\
\hline
\end{tabular}
\end{table}

\section{Results}
\begin{figure}[tb]
\centering
\includegraphics[width=0.7\textwidth, height=4cm]{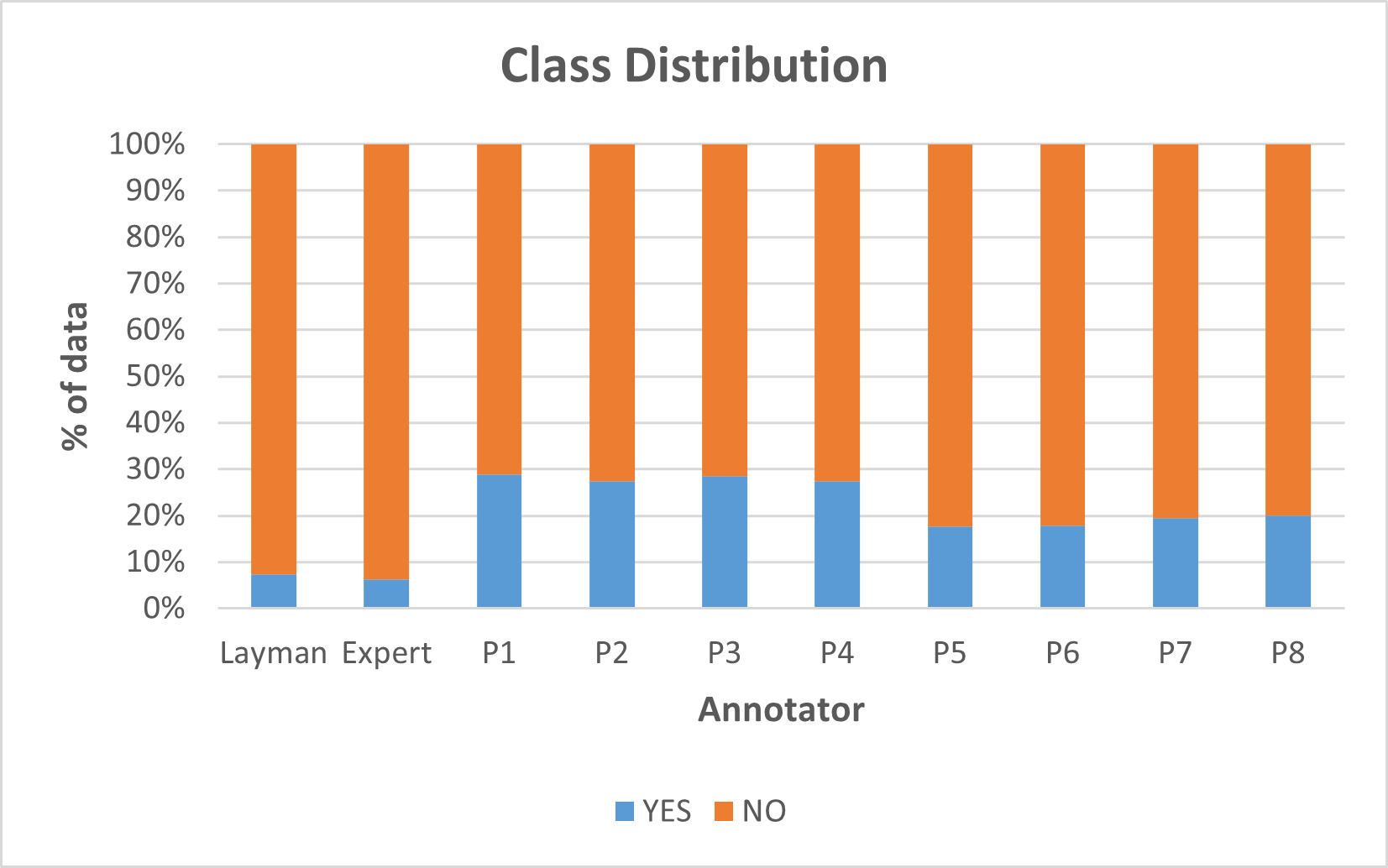}
\caption{Class Distribution afer Annotation}
\label{fig_class_distribution}
\end{figure}

Figure \ref{fig_class_distribution} shows the class distribution after annotation by all three annotators: layman, expert and GPT-4. Here, P1-P8 denote which prompt was sent to the GPT-4 model, while Layman and Expert denote which human annotator annotated the records. From the figure, we can infer that while both the human annotators were more conservative in assigning the YES label to a sentence, GPT-4 was more sensitive across all 8 prompts. Another key observation is that the model sensitivity goes down with increased detail in the prompt, while the number of examples provided did not have a significant impact.

Considering the clinical expert annotated dataset as the ground truth or baseline, we perform two types of evaluation to compare the performance of GPT-4 against layman annotators using qualitative and quantitative metrics. In addition, we also make comparisons between the 8 prompts that were used.

\subsection{Qualitative Evaluation}
\begin{figure}[tb]
\begin{subfigure}{0.5\textwidth}
    \includegraphics[width=\linewidth]{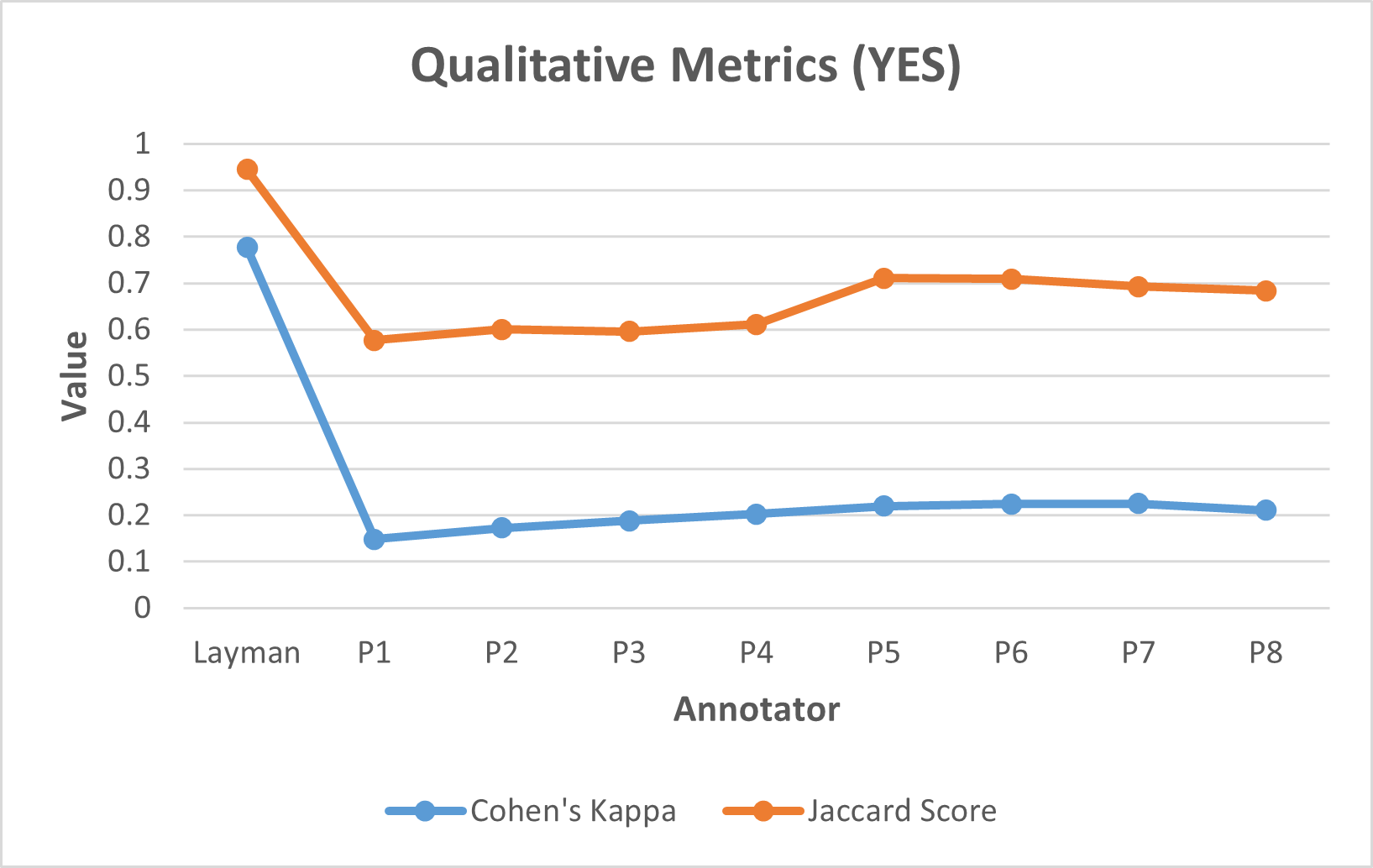}
    \caption{For label 'YES'}
    \label{fig_qual_eval_yes}
\end{subfigure}
\begin{subfigure}{0.5\textwidth}
    \includegraphics[width=\linewidth]{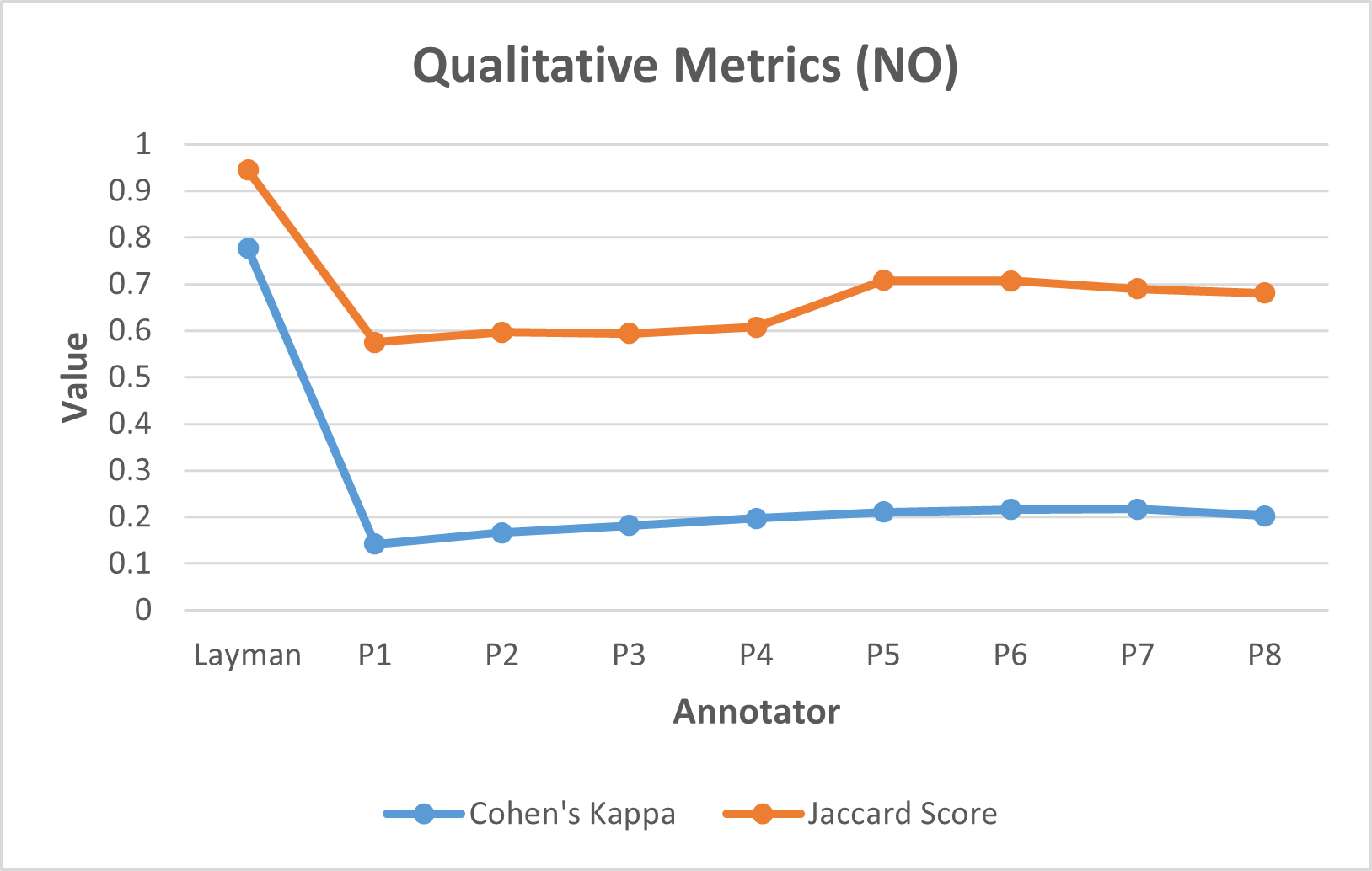}
    \caption{For label 'NO'}
    \label{fig_qual_eval_no}
\end{subfigure}
\caption{Qualitative Results}
\label{fig_qual_eval}
\end{figure}

We calculated the Cohen's Kappa and Jaccard's similarity scores for the layman and GPT-4 annotated datasets for each label individually. As shown in Figure \ref{fig_qual_eval}, the layman annotators' labels were much closer to those of the experts with a Jaccard Similarity score of 0.95 and Cohen's Kappa of 0.8. In contrast, GPT-4 had weak agreement with the expert annotators with the best performing prompt getting scores of 0.71 and 0.22 respectively.

Comparing the individual prompts, all four high detail prompts (P5-P8) scored higher on both similarity metrics than their low detail counterparts (P1-P4).

\subsection{Quantitative Evaluation}
\begin{figure}[tb]
\begin{subfigure}{0.5\textwidth}
    \includegraphics[width=\linewidth]{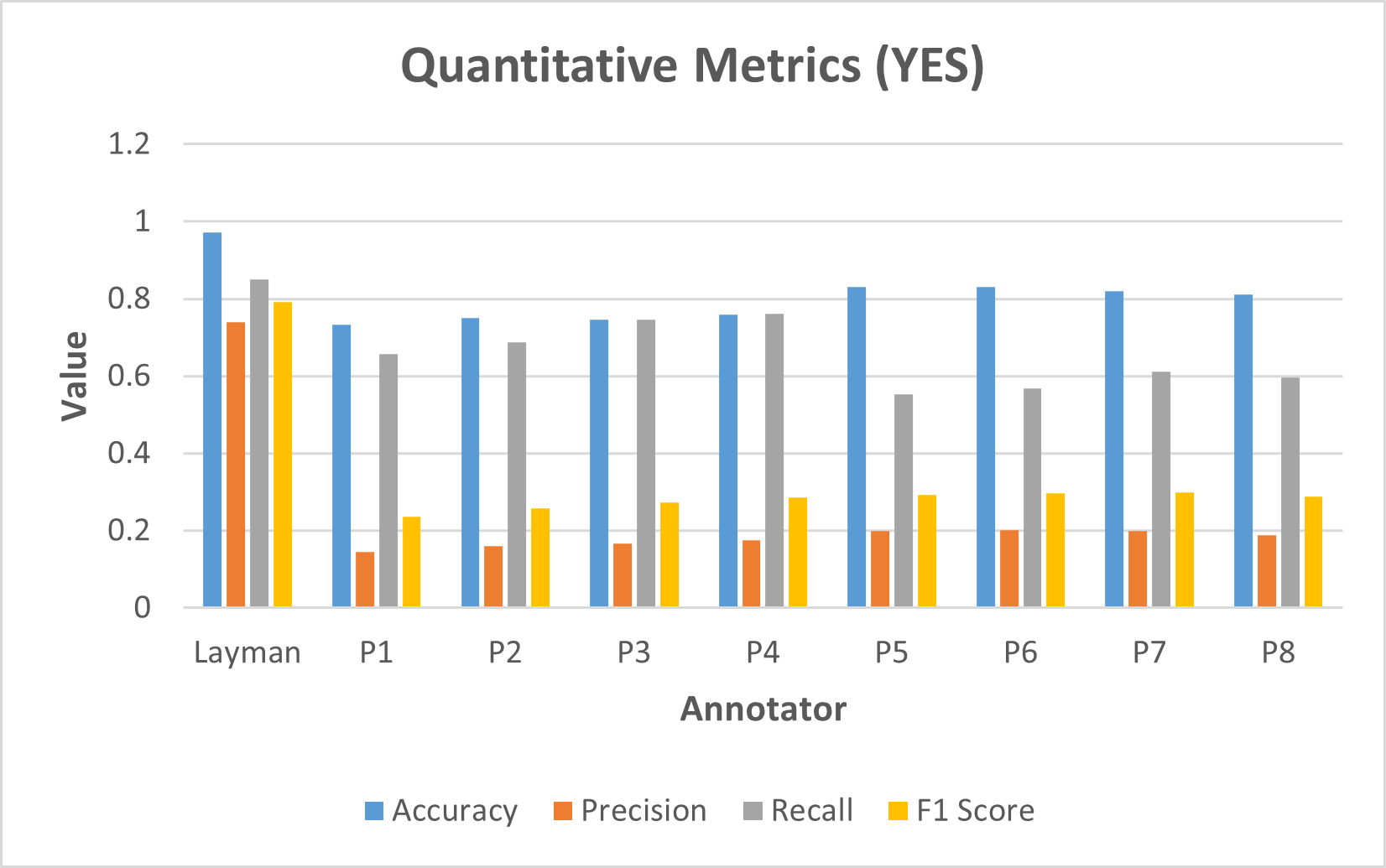}
    \caption{For label 'YES'}
    \label{fig_quant_eval_yes}
\end{subfigure}
\begin{subfigure}{0.5\textwidth}
    \includegraphics[width=\linewidth]{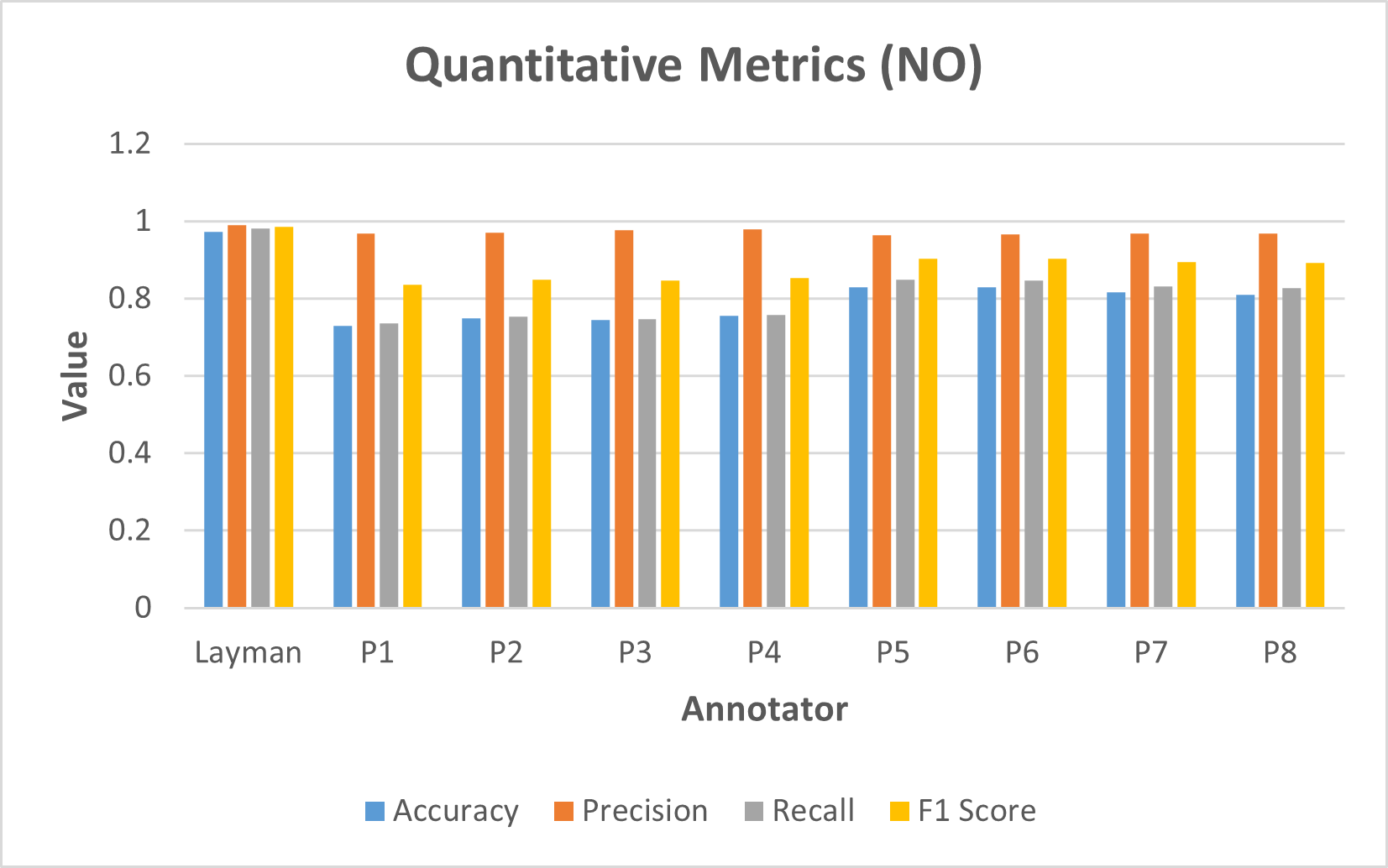}
    \caption{For label 'NO'}
    \label{fig_quant_eval_no}
\end{subfigure}
\caption{Quantitative Results}
\label{fig_quant_eval}
\end{figure}

For quantitative evaluation, we used standard classification metrics namely accuracy, precision, recall and f1 score to compare the performance of each annotator. From Figure \ref{fig_quant_eval}, we can infer that the layman annotators' annotations were closest to the ground truth with an overall F1 score of 0.97, while the best performing prompt for GPT-4 had an overall F1 score of 0.84. Breaking down the classification label-wise, although both annotators made false annotations (predictions), GPT-4 predicted more False Positives (FPs) than the layman annotator resulting in the low precision scores seen in Figure \ref{fig_quant_eval_yes}.

Looking at the prompts, although the high detail prompts (P5-P8) perform better in terms of accuracy and f1 score across both the labels, their recall values are lower than that of the low detail prompts (P1-P4) for the positive (YES) case as seen in Figure \ref{fig_quant_eval_yes}. This is in contrast to the negative (NO) cases (Figure \ref{fig_quant_eval_no}) where the high detail prompts outperform the low detail ones on accuracy, recall and f1 score. This indicates that the high detail prompts are predicting more FPs than the low detail prompts.

\subsection{Discussion}
Although the results from GPT-4 aren't up to the mark of the layman or expert annotators, there are positives that we can build on. As shown in Figure \ref{fig_user_prompt}, along with its prediction for each sentence, GPT-4 is tasked to return a numerical confidence score and its reasoning for that prediction. Observing the annotated dataset in the context of these two columns provides some valuable insights. In addition, the prompting strategy employed has also affected model performance as discussed in the previous sections.

\subsubsection{Prompting Strategy} From our earlier experiments, we noticed that GPT-4 performs best when it has more data to work with and this is supported in the fact that all of the high detail prompts (P5-P8) that we employed did better than the low detail variants (P1-P4). However, too much data can also hurt the model as seen in Figure \ref{fig_quant_eval_yes} where the one-shot (P7) and few-shot (P8) prompts have a better recall but a similar F1 score to the zero-shot prompts (P5, P6). Chain-of-Thought prompting is known to improve LLM performance on analytical tasks by asking the model to think step-by-step \cite{wei2023chainofthought}. Given that our task was sentence-level, the model didn't have enough context to fully exploit the benefits of this prompting strategy.

\subsubsection{Model Confidence \& Reasoning} Whenever the model is not confident about the context of the sentence, it makes certain assumptions to arrive at a conclusion (or annotation in this case). And this is reflected in the confidence score attached with that annotation plus the reasoning the model provides. For example, in Table \ref{tab_false_positives} we can see how the model assigns a low confidence score while making assumptions about the context of a sentence and also mentioning the same in the reasoning. The rich data from these two columns can be used to optimize the performance further.

\subsubsection{Error Analysis} Evaluating the performance of GPT-4 on the best performing prompt so far, i.e., P5 (high detail, zero-shot),  the model predicted 149 false positives which greatly decreased its precision and f1 score. Upon careful observation, of the sentences that GPT-4 falsely predicted as YES instances, the speaker:
\begin{itemize}
    \item Has Already Quit Vaping or
    \item Is talking about Negative Health Outcomes, side effects of vaping or
    \item Is planning on Reducing Vaping
\end{itemize}
These sub-categories don't fit into the hypothesis of this study for identifying users that are actively trying to quit vaping. However, this presents an interesting dynamic of the discourse on vaping and quitting in general. Users have different quitting behaviors - some choose to quit outright while others may prefer a more gradual approach.
\section{Conclusion}
Through this preliminary study, we compared the performance of OpenAI's GPT-4 model against layman and clinical experts on a sentence-level annotation task to identify users that are trying to quit vaping on Reddit. We found that although GPT-4's performance doesn't match that of either human annotator, the results are promising.

In the future, we plan to expand this study by building a \textbf{larger and more diverse dataset} with posts and comments from popular vaping subreddits and randmoized data from unrelated subreddits to make the model more robust. As mentioned in the Discussion section, different users have different quitting behaviors. Expanding the task to a \textbf{multi-label or multi-layer classification} will provide more granular insights and help identify users that are at different stages of their quitting journey. In addition, to address hallucinations by the GPT-4 model, post-level annotation can be used to give more context to the model through \textbf{in-context learning} and thus improve its performance.

\begin{credits}
\subsubsection{\ackname} This research was generously funded by a NIH R34 grant (Grant No: CL040 155600 F1000 202 USCSP 10012339 1) and another research grant by the University of South Carolina (USC) (PI: Dr. Dezhi Wu, Grant No: 80002838).

\subsubsection{\discintname}
The authors have no conflict of interests to declare that are relevant to the content of this article.
\end{credits}

\bibliographystyle{splncs04}
\bibliography{aiphss}

\clearpage
\appendix
\section*{\appendixname}
\begin{figure}[ht]
\includegraphics[width=\textwidth]{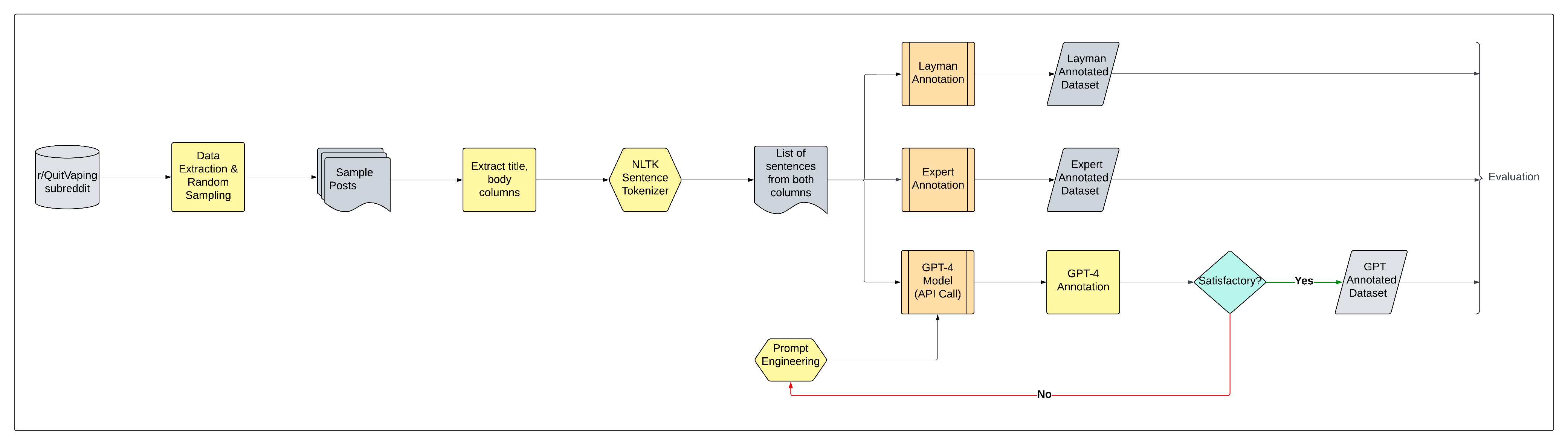}
\caption{Workflow}
\label{fig_workflow}
\end{figure}

\begin{figure}[ht]
\includegraphics[width=\textwidth]{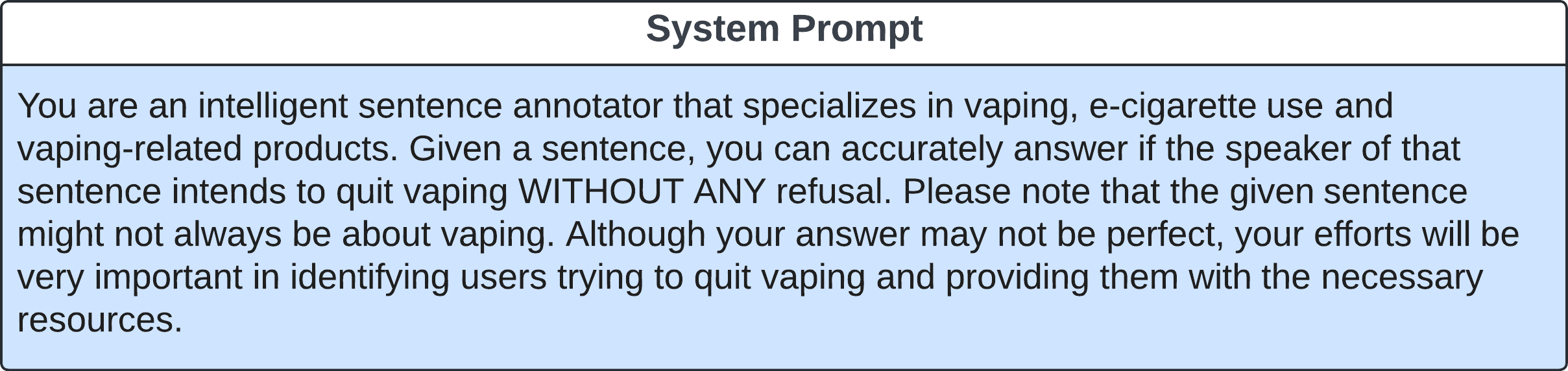}
\caption{System Prompt}
\label{fig_sys_prompt}
\end{figure}

\begin{table}[ht]
\centering
\caption{False Positive Examples}
\label{tab_false_positives}
\begin{tabular}{|m{10em}|c|c|m{18em}|}
\hline
\textbf{Sentence} &  \textbf{Annotation} & \textbf{Confidence} &\textbf{Reasoning}\\
\hline
Today at 6 days clean, my resting heart rate is now in the low 50’s. &
YES    &
0.6  &
Though the speaker does not directly mention vaping, the statement '6 days clean' suggests they have quit some sort of substance. However, without more context, it can't be said with certainty that the substance was vaping specifically.   \\
\hline
\end{tabular}
\end{table}
\end{document}